\documentclass[lettersize,journal]{IEEEtran}
\usepackage{amsmath,amsfonts}
\usepackage{algorithmic}
\usepackage{algorithm}
\usepackage{array}
\usepackage[caption=false,font=normalsize,labelfont=sf,textfont=sf]{subfig}
\usepackage{textcomp}
\usepackage{stfloats}
\usepackage{url}
\usepackage{verbatim}
\usepackage{graphicx}
\usepackage{cite}
\begin{document}

\title{Dynamic Texture Transfer using\\ PatchMatch and Transformers}

\author{Guo Pu, Shiyao Xu, Xixin Cao, and Zhouhui Lian~\IEEEmembership{Member,~IEEE,}
\thanks{Guo Pu, Shiyao Xu and Zhouhui Lian are with Wangxuan Institute of Computer Technology, Peking University, Beijing {\rm 100871}, China. Xixin Cao is with School of Software and Microelectronics, Peking University, Beijing {\rm 100871}, China}
\thanks{Corresponding Author: Zhouhui Lian (email: lianzhouhui@pku.edu.cn)}
}



\maketitle

\begin{abstract}
How to automatically transfer the dynamic texture of a given video to the target still image is a challenging and ongoing problem. In this paper, we propose to handle this task via a simple yet effective model that utilizes both PatchMatch and Transformers. The key idea is to decompose the task of dynamic texture transfer into two stages, where the start frame of the target video with the desired dynamic texture is synthesized in the first stage via a distance map guided texture transfer module based on the PatchMatch algorithm. Then, in the second stage, the synthesized image is decomposed into structure-agnostic patches, according to which their corresponding subsequent patches can be predicted by exploiting the powerful capability of Transformers equipped with VQ-VAE for processing long discrete sequences. After getting all those patches, we apply a Gaussian weighted average merging strategy to smoothly assemble them into each frame of the target stylized video. Experimental results demonstrate the effectiveness and superiority of the proposed method in dynamic texture transfer compared to the state of the art.
\end{abstract}


\section{Introduction}
\IEEEPARstart{N}{owadays},  with the rapid development of computer and network technologies, dynamic textures have become popular in a variety of media, such as films, digital posters, advertisements, and online chatting, making contents more appealing and varied. However, designing a set of dynamic special effects takes numerous times and efforts even for skilled designers. Taking artistic dynamic texts as an example, font designers need to design the special text effect frame by frame, and apply this dynamic effect to different characters in the entire font library, which is extremely labor-intensive and time-consuming. Hence, the aim of this paper is to automatically transfer the dynamic effect of a given video sample to a target shape or image, significantly improving the designing efficiency of dynamic textures.

\begin{figure}[t]
    \centering
    \includegraphics[width=1\linewidth]{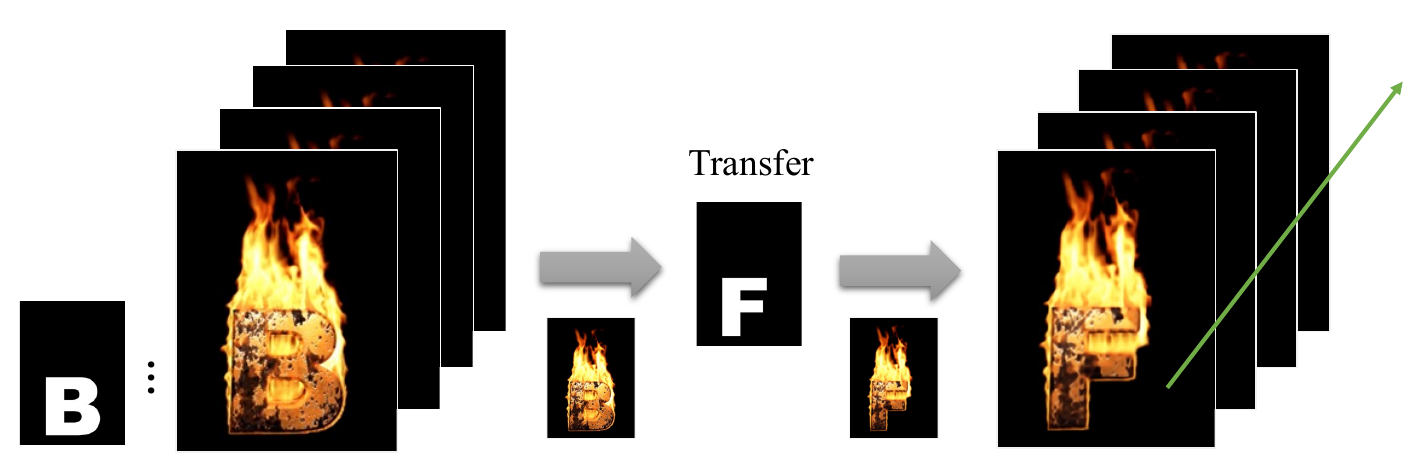}
    \caption{An example of dynamic texture transfer. Given a sample video and a target still image, the proposed method is able to synthesize the target video by transferring the dynamic texture of the sample video into the target image.}
    \label{Overview}
\end{figure}

The above-mentioned process is called dynamic texture transfer. Compared with static texture transfer, both the internal spatial information of the image and the temporal information across all frames need to be considered. What is more, the flame flickering and shaking problems that appear in the video synthesis process also need to be avoided. 

Most current texture transfer methods~\cite{yang2017awesome},~\cite{men2018common},~\cite{yang2018context},~\cite{shaham2019singan} aim at handling static images which do not apply temporal constraints thus fail to satisfactorily address the dynamic texture transfer problem. Moreover, existing neural-based methods~\cite{xie2017synthesizing},~\cite{tesfaldet2018two},~\cite{siarohin2019first},~\cite{thomas2020learning} typically require huge amounts of data to train their models which are incapable of handling the one-shot effects transfer task. Furthermore, these neural-based models are required to resize all images to a certain aligned scale during training which causes detail losing of delicate textures, while our method works in a full-resolution fashion and thus is capable of transferring complex textures on their original resolution. The most relevant work to our paper is~\cite{men2019dyntypo}, where Men et al.~\cite{men2019dyntypo} proposed to stylize a still semantic map to the target video. However, since a common Nearest-neighbor Field (NNF) is adopted to guide the texture synthesis procedure across all keyframes simultaneously, their method fails to apply to complex dynamic effects (e.g. with motions). In addition, the correlation of video frames is not properly exploited and thus there still exist some incoherent regions in their synthesis results. 

To solve the above issues, we propose DynTexture, a neural-based approach to automatically transfer various dynamic effects. As shown in Fig.~\ref{Overview}, our method requires only a source texture video and a target semantic map as input, then it can automatically generate the target video with the same texture effect as the source video.

To the best of our knowledge, the proposed method is the first neural-based approach to tackle this one-shot dynamic texture transfer problem. To animate the binary semantic image, we first introduce an attention term called distance map, which can help to better preserve the structural information. With the guidance of distance maps, the offset and the correspondence between inputs are calculated and the initial target frame can then be synthesized via the PatchMatch~\cite{barnes2009patchmatch} algorithm. Afterwards, we cut the start frame into structure-agnostic patches and calculate their embeddings to make better use of the texture continuity and the patch distributions inside the image. In the second stage, we use the Transformer~\cite{vaswani2017attention} model to predict the discretized code sequences, which is capable of capturing the long-distance dependencies between frames.

In summary, our major contributions are threefold:
\begin{itemize}
    \item We obtain impressive results in the dynamic text effects transfer task with a new texture transfer and animation scheme that consists of a distance map guided texture transfer module and a novel deep sequence forecasting module. Extensive experiments demonstrate our method's superiority over state-of-the-art methods and versatility in various texture transfer and animation tasks.
    \item We tackle the challenge of insufficient data when training the deep sequence forecasting module through overlapping patch splitting and merging data augmentation, enabling the model to effectively resolve the one-shot dynamic texture transfer problem. 
    \item The independence of the texture rendering module and the sequence forecasting module makes our method capable of synthesizing more complex dynamic effects (e.g., with motions) that cannot be properly handled by existing state-of-the-art approaches (e.g.,~\cite{men2019dyntypo}).
\end{itemize}

\section{Related Work}

Dynamic texture transfer has been a challenging task. At present, there exist very few works that try to solve this problem. In this section, we briefly survey the texture transfer and animating literature and summarize the current state of the art. The following highlights the techniques most closely related to ours.

The still texture effects transfer task has been well resolved by existing approaches (e.g.,~\cite{yang2017awesome},~\cite{men2018common},~\cite{yang2018context},~\cite{yang2019controllable}) that take the input semantic map as structure guidance to learn the correspondence between the source and target images. Yang et al.~\cite{yang2017awesome} proposed to transfer the text effects by analyzing the high regularity of the spatial distribution for target effects. Actually their work can be seen as the extension of \cite{barnes2009patchmatch}, \cite{barnes2010generalized}. They performed transferring by calculating the similarity attractively between patches to find the best matching. As for deep-learning-based methods, Yang et al.~\cite{yang2019tet} designed a GAN-based network to accomplish both the objective of style transfer and style removal so that it can learn to disentangle and recombine the texture of content. However, like most deep learning models, their work requires a lot of paired data and is unsuited to handle the one-shot texture transfer task in our case. Shaham et al.~\cite{shaham2019singan} trained the model only on one single image to capture the internal distribution of patches, using a pyramid of fully convolutional GANs, to capture details at different scales. However, when applying the method~\cite{shaham2019singan} on complex cases, for example, the burning flame with dynamic texture effects, the generated results are blurry and can hardly preserve the desired structure. Since temporal consistency is not considered, the above-mentioned methods fail to handle the task of dynamic effects transfer. 

To take temporal constraints into consideration, dynamic effects transfer methods like~\cite{bhat2004flow} perform dynamic effects transfer by analyzing the motion of textured particles in the video while the method is not a fully automatic method due to the requirement of user-specified flows. Tesfaldet et al.~\cite{tesfaldet2018two} utilized a pre-trained convolutional network to model the dynamics of per-frame appearances of textures to generate target textures. Xie et al.~\cite{xie2017synthesizing} defined a probability distribution over video sequences and introduced the temporal constraint through multi-scale model layers to capture spatial-temporal patterns. However, these methods require plenty of training data and struggle to capture the details of textures, thus failing to deal with the one-shot texture transfer task.

Most image animation approaches~\cite{benard2013stylizing},~\cite{fivser2017example},~\cite{men2019dyntypo},~\cite{thomas2020learning},~\cite{siarohin2019first} either require large amounts of videos for training, failing to work in the one-shot learning fashion, or have strong restrictions for the input objects such as human poses, thus cannot be applied to synthesize dynamic textures. 

Recently, Men et al.~\cite{men2019dyntypo} extended the NNF search of PatchMatch~\cite{barnes2009patchmatch} to the spatial-temporal domain and obtained impressive results in dynamic text effects transfer. They restricted the procedure of patch searching within the extracted keyframes to maintain both spatial and temporal consistencies, in the meantime losing inter-frame information which causes incoherent structures in their synthesis results. Moreover, their method only searches for a single global NNF, and thus fails to process dynamic textures with moving samples which shift across source videos.

Transformer-based methods are emerging in Computer Vision tasks, demonstrating their inherent advantages in capturing relationships and dependencies between sequences. Esser et al.~\cite{esser2020taming} introduced the convolutional VQ-VAE~\cite{oord2017neural} to learn compressed discrete representations for images and then modeled learnt embeddings with Transformers to generate controllable realistic images. Yan et al.~\cite{yan2021videogpt} used a simple GPT-like architecture to model the discrete latent codes learnt from videos by VQ-VAE~\cite{oord2017neural} to generate new videos. Constrained by the compressing ability of VQ-VAE, their results are limited to a very low resolution.

\begin{figure}[t]
    \centering
    \includegraphics[width=0.5\textwidth]{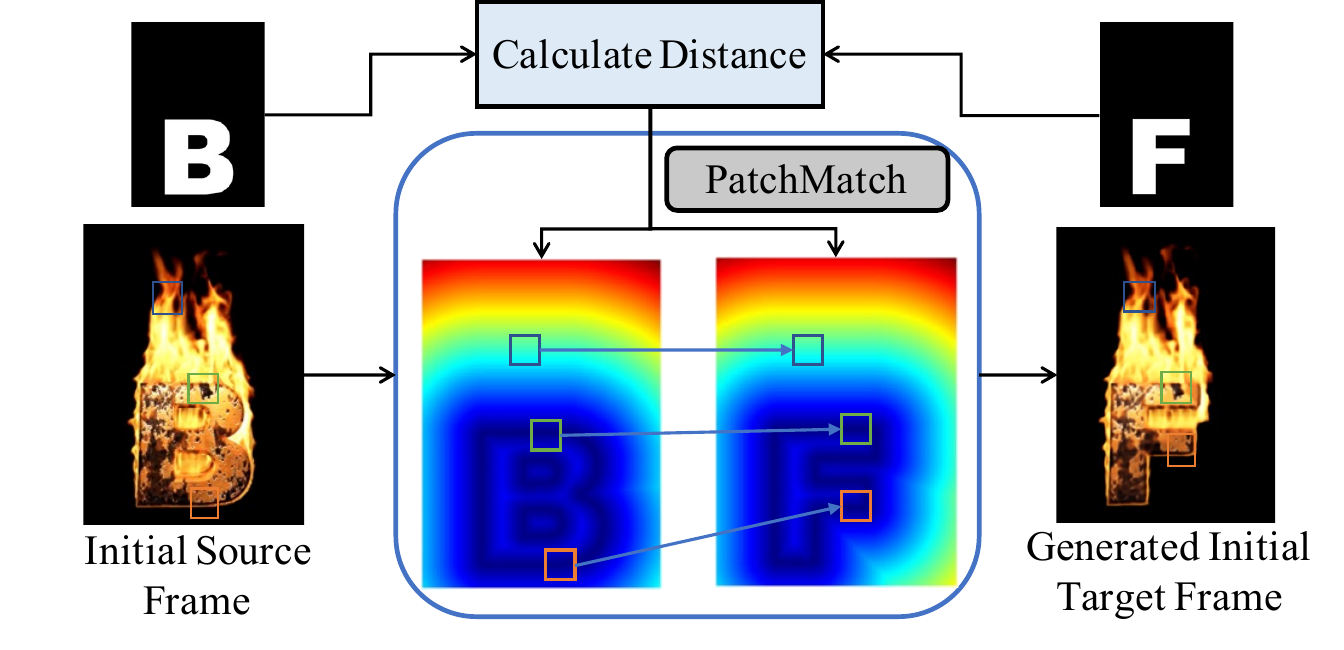}
    \caption{Utilizing distance information to guide the PatchMatch algorithm, letting the flow of information outward from the boundary.}
    \label{fig:3}
\end{figure}

\section{Method Description}

In this section, we first formulate the task of dynamic texture transfer (Section 3.1) and then present the detailed method description of our approach (Section 3.2 and 3.3), which is named as DynTexture (Dynamic Texture transfer) for short.

\begin{figure*}[t!]
    \centering
    \includegraphics[width=0.8\linewidth]{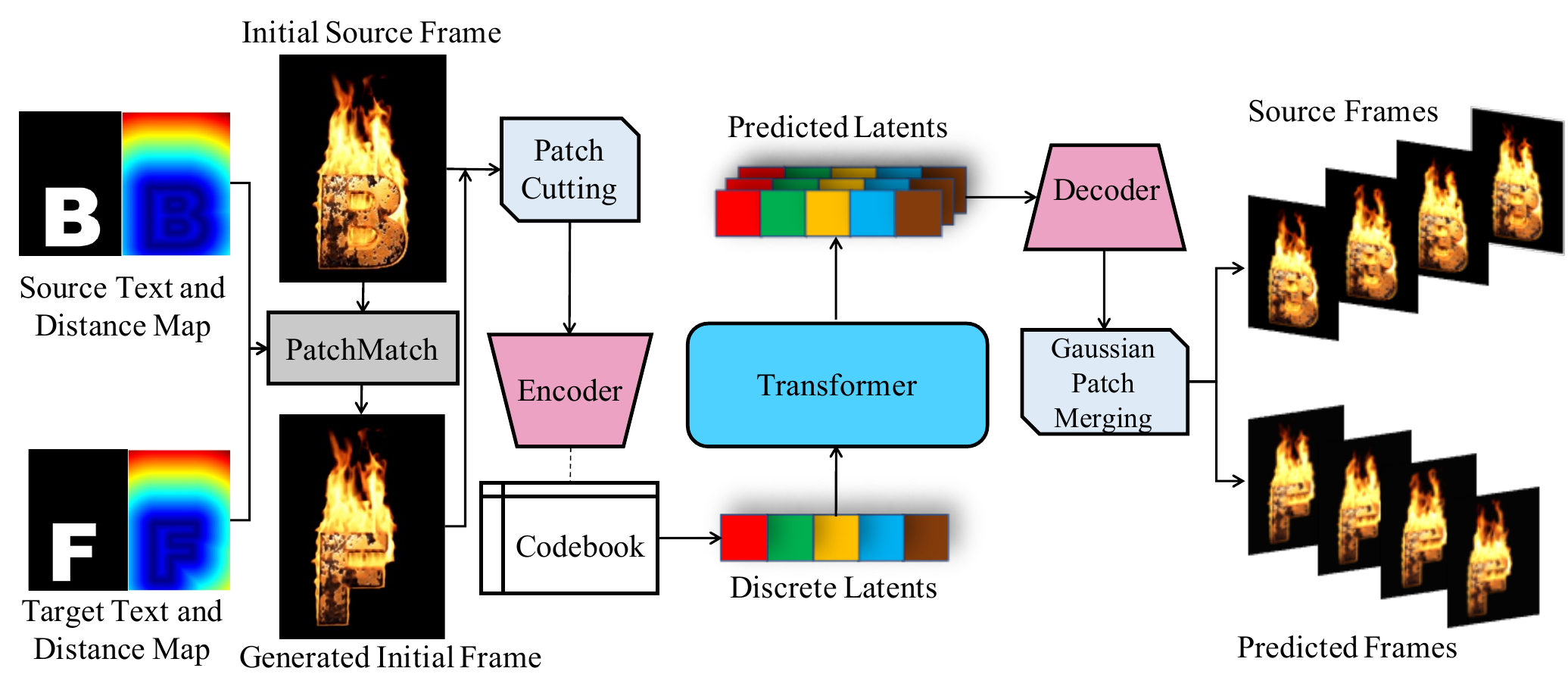}
    \caption{Overview of the proposed DynTexture, which is designed as a two-stage architecture, where the distance map guided texture rendering module generates the initial frame, and the novel deep sequence forecasting module predicts and synthesizes the subsequent frames based on the previously-synthesized initial frame.}
    \label{fig:4}
\end{figure*}

Fig.~\ref{fig:4} demonstrates an overview of our proposed DynTexture. The key idea is to decompose the one-shot dynamic texture transfer task into two stages where the start frame of the target video with the desired dynamic texture is synthesized in the first stage (Section 3.2) via a distance map guided texture transfer module based on the PatchMatch algorithm~\cite{barnes2009patchmatch}. Then, in the second stage (Section 3.3), the synthesized image is decomposed into structure-agnostic patches, according to which their corresponding subsequent patches can be predicted by exploiting the powerful capability of Transformers~\cite{vaswani2017attention} equipped with VQ-VAE~\cite{oord2017neural} for processing long discrete sequences. After getting all those patches, we apply a Gaussian weighted average merging strategy to smoothly assemble them into each frame of the target stylized video.

\subsection{Problem formulation and analysis}
The proposed DynTexture can be used to handle various types of dynamic texture transferring tasks. For the sake of convenience, we formulate the task of dynamic texture transfer under the scenario of dynamic text effects transfer. More specifically, given a source text image $S_{text}$, its stylized animation $S_{style}$ and a target image $T_{text}$, the goal of dynamic text effects transfer is to synthesize the target stylized animation $T_{style}$ such that $S_{text}$:$S_{style}$::$T_{text}$:$T_{style}$, as shown in Fig.~\ref{Overview}. We decompose the task of dynamic text effects transfer into two sub-problems: still image texture transfer and dynamic effects animation. 
    For the first sub-problem, PatchMatch-based methods like~\cite{yang2017awesome} have achieved impressive results for texture transfer in still images. We tackle the still image texture transfer based on PatchMatch as well. 
    As for how to transfer the dynamic effects to the still target image, we take advantage of the Transformers' long sequence prediction ability, decomposing the image into large amounts of structure-agnostic patches and training Transformers to learn the correlation between frames. 

\subsection{Initial frame synthesis}

We use the distance map guided PatchMatch~\cite{barnes2009patchmatch} algorithm to search an NNF for the initial frame and generate the initial frame with the desired texture effect. 

For still text effects transfer, the target stylized image is synthesized with the semantic guidance $T_{text}$. Obviously, the patches in the text contour contain more features and could easily find proper correspondences from the source. By exploiting the preserving and propagating features of the PatchMatch algorithm, we can find the optimal matches by enforcing texture continuity within and near the contour. This guarantees the spatial consistency of the initial frame.

Unfortunately, the naive utilization of PatchMatch~\cite{barnes2009patchmatch} in the application of text effects transfer fails to transfer complicated textures without semantic guidance outside the text (see the second row of Fig.~\ref{fig:10}).
Inspired by~\cite{men2019dyntypo}, we calculate a distance map as the propagation guidance to enable the texture effect to spread around the text contour. Following~\cite{men2019dyntypo}, the distance is calculated as the normalized distance between each pixel of $T_{text}$ and the contour of text. The initial frame synthesis process is illustrated in Fig.~\ref{fig:3}.

\subsection{Subsequent frame prediction}
\subsubsection{Patch cutting}

The source exemplar contains both source structure information and source texture effects where we only want the model to learn the texture information and ignore the other.
	To achieve that, we cut overlapping patches for every frame in the source video and the target image, and process them independently in later procedures.
    
	The purpose of patch cutting is to decompose the input image into structure-agnostic patches. We can also consider this as a way of data augmentation as it produces huge amounts of training data for the model.
	The patches are overlapped since we want to make the model be aware of the continuity between its neighbours and encourage the model to generate smooth results. Consequently, a smaller patch cutting stride leads to better results and the stride is set to 1 in our method. 
     
     The choice of patch size is crucial for patch-based synthesis methods like the proposed DynTexture. A smaller patch size makes the patches less discriminative and causes a large portion of repetitive patterns or even misplaced patterns. On the contrary, a large patch size tends to contain too much information of the source structure and thus affects the target structure of the synthesized video, which is unsatisfactory. We experimentally choose the patch size as 16 by 16 which results in a good trade-off between better patch discrimination and less structural damage.

\subsubsection{Compressing patches by VQ-VAE}

Using Transformers to predict long subsequent patch sequences requires us to learn a rich codebook to effectively compress patches.
VQ-VAE proposed in~\cite{oord2017neural} is able to reconstruct sufficiently realistic images by representing them with its codebook entry codes.
Thereby, we employ VQ-VAE to learn the texture patches from the source exemplar and compress the patches into compact latent codes (i.e., code-book entry indices) for further processing. To make it adaptive for the small patch input, we slightly modify the original VQ-VAE model to a small-scale version, which is trained by computing the reconstruction loss, the codebook loss, and the commitment loss:
\begin{equation}
\label{eqn:01}
\begin{aligned}
L_{VQVAE} = ||x-D(e)||^{2}_{2} + ||sg[E(x)]-e||^{2}_{2} + \\ 
\beta ||sg[e]-E(x)||^{2}_{2}, \\
\end{aligned}
\end{equation}
where $E$ and $D$ are the encoder and the decoder of VQ-VAE, respectively, $x$ is the input patch, $e$ is the latent embedding of $x$, $\beta$ is the coefficient of the commitment loss, and $sg$ denotes the stop-gradient operation. The objective loss function $L_{VQVAE}$ consists of three terms where the first term is the reconstruction loss that supervises VQ-VAE to learn good representations to reconstruct the patches. The second term and the third term are the codebook loss and the commitment loss, respectively, guiding the model to learn a good codebook.

The encoder and the decoder of VQ-VAE are trained on the patches from source frames, thus the patches can be represented in terms of the codebook indices of their encodings. Specifically, given a patch $x$, we denote the codebook indices of $x$ as $I_{x}$. Elements in $I_x$ are the nearest entries among all elements in the codebook with respect to $Z_{q}$, and can be formerly calculated as:
\begin{equation}
\label{eqn:01}
I_{x_{ij}} = \mathop{\arg\min}_{\theta \in \Theta }  Dist(Z_{q_{ij}}, Z_{\theta}),
\end{equation}

\begin{equation}
\label{eqn:01}
Dist(Z_{q_{ij}}, Z_{\theta}) = \left \| Z_{q_{ij}} - Z_{\theta} \right \|_2,
\end{equation}
where $Z_{q}$ is the quantized encoding of $x$, $Z_{\theta}$ is the $\theta$-th element in the codebook, $\Theta$ denotes the size of codebook, $Dist$ denotes the distance calculation function, and $i$ and $j$ denote the indices of element locations.

Notice that the patch $x$ can be easily recovered by firstly mapping $I_x$ to $Z_q$, and then decoding $Z_q$ via the decoder. We will recover the predicted subsequent patch index sequence in the same manner.

Utilizing VQ-VAE to encode the patches into latent codes brings two benefits:
	First, VQ-VAE learns the texture information from structure-agnostic patches and represents the patches as high-quality latent codes, and it is able to generate new patches through manipulations on the latent codes.
 	Second, VQ-VAE achieves a high compressing rate by converting patch images to latent codes, which markedly reduces the prediction burden of the deep sequence forecasting module, and thus enables long sequence prediction with limited computing and memory resources.
 	
\subsubsection{Subsequent sequence prediction via Transformers}

Transformers~\cite{vaswani2017attention} have shown tremendous successes in modeling discrete data such as natural languages. The architecture of Transformers employs multi-head self-attention blocks followed by point-wise MLP feed-forward blocks. It is particularly suitable for the generation of long discrete sequences which coincides with our desire for predicting the discrete latent codes of subsequent sequences.
	After encoding the initial frame patches into discrete latent codes, we utilize Transformer to predict the subsequent latent code sequences.  
	Specifically, the indices of initial frame patches are fed into the Transformer model and then the predicted indices of the subsequent patch sequence are outputted. During the training stage, we concatenate the patch indices of the subsequent sequence into a long sequence as the target sequence. Formally, the Transformer model used here aims to fit the following function:
    \begin{equation}
\label{eqn:01}
    P(I_{0}, I_{1}, ..., I_{F}|I_{0}) = \prod_{i=1}^{F}P(I_{i}|I_{0}),
    \end{equation}
   	where $P$ denotes the probability distribution, $I_{i}$ denotes the patch index of the i-th frame, and $F$ denotes the total number of frames.
    
    The cross-entropy loss is used to train the model:
\begin{equation}
\label{eqn:01}
L_{Transformer} = -\sum_{}^{}p(I_{pred})\log q(I_{GT}),
\end{equation}
where $I_{pred}$ and $I_{GT}$ are the predicted and ground-truth patch indices, respectively, whose distributions are denoted by $p(I_{pred})$ and $q(I_{GT})$, respectively.

    Once our model has learnt the correspondence between the indices of the initial frame and those of the subsequent frames, it will ensure the temporal consistency of the decoded video. As a result, each synthesized image patch will imitate the appearance variance learnt from the source patches while acquiring motion properties implicitly.

\subsubsection{Patch decoding and merging}

\begin{figure}[!t]
    \centering
    \includegraphics[width=0.4\textwidth]{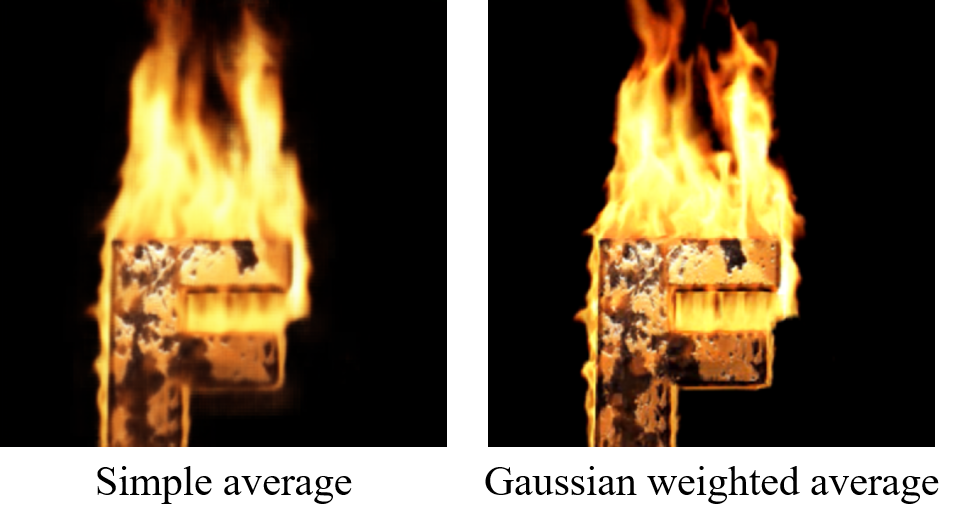}
    \caption{Comparison between simple average and Gaussian weighted average merging strategies. Gaussian weighted average obviously obtains higher-quality results.}
    \label{fig:5}
\end{figure}

After getting the predicted patch indices in the subsequent sequence, we use the trained decoder of VQ-VAE to decode the predicted patch indices and assign them to the corresponding frames.
	
Since there exist overlapping regions between adjacent patches, a good patch merging strategy is crucial for patch-based image synthesis. As shown in  Fig.~\ref{fig:5}, applying simple average to the merging process causes blurry effects. On the contrary, Gaussian weighted average obtains decent results so that we utilize it to merge these patches to generate the output frames. We experimentally set the standard deviation of the Gaussian kernel, which affects the sharpness of the merged image, to 4 in our method to obtain the best performance. More specifically, the value of a pixel in the synthesized image is computed as:
\begin{equation}
\label{eqn:01}
    P_{i_{xy}} = \frac{\sum_{p_{xy}> 0 }^{} w_{p_{xy}}p_{xy}}{\sum_{p_{xy}> 0 }^{}w_{p_{xy}}},
\end{equation}
where $i=1,{\ldots}\,,T$ denotes the index of a frame, $p_{xy}$ means the value of the pixel ($x$,$y$) in the patch $p$, and $w_{p_{xy}}$ denotes the pixel's Gaussian weights and factors, which is defined as:
\begin{equation}
\label{eqn:01}
w_{p_{xy}} = \frac{1}{2\pi \sigma ^{2}}e^{-\frac{(x-m)^{2} + (y-n)^{2}}{2\sigma^{2} }},
\end{equation}
where $\sigma$ is the standard deviation of the Gaussian kernel and ($m$,$n$) denotes the center of the patch $p$.

\section{Experiments}

Our approach is initially designed for dynamic text effects transfer, but it can also be applied to other texture transfer applications. In this section, we first evaluate the effectiveness of our approach in handling the task of dynamic text effects transfer with various font styles and several representative types of glyphs (e.g., English and Chinese characters), and illustrate its superiority over state-of-the-art methods. Then, we further demonstrate the capability of our method in other applications such as image animation.

\subsection{Dynamic text effects transfer}

We apply our dynamic texture transfer approach to transfer the dynamic text effects of many stylized examples to the glyph images of English and Chinese characters. We demonstrate that various complicated dynamic effects such as burning flame, flowing water, and others can be accurately transferred by our approach. Some results are shown in Fig.~\ref{fig:6} and~\ref{fig:7}, and more can be found in the supplemental material.

\begin{figure}[t]
    \centering
    \includegraphics[width=0.5\textwidth]{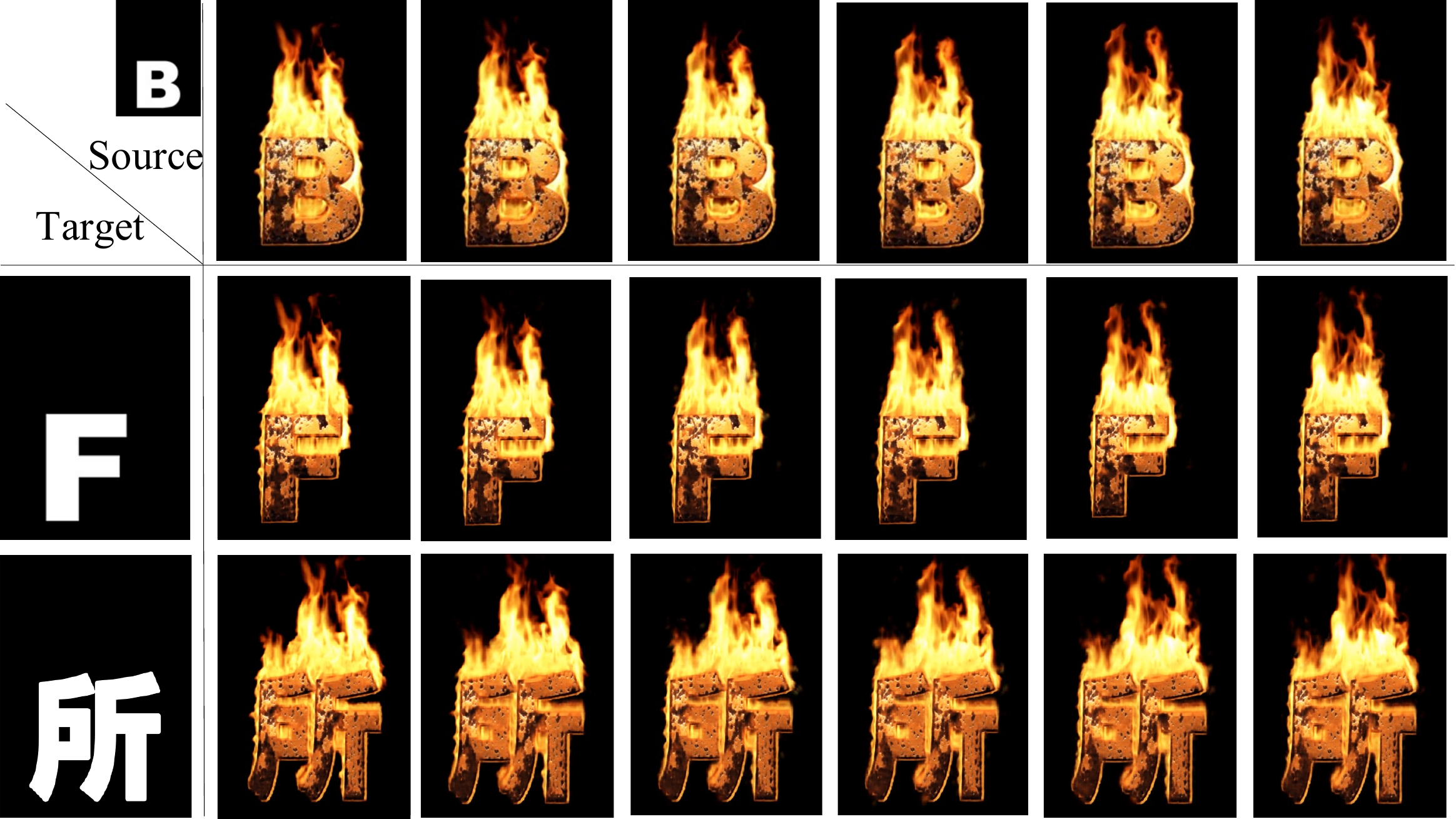}
    \caption{Our results on flame dynamic effects transfer.}
    \label{fig:6}
\end{figure}
\begin{figure}[t]
    \centering
    \includegraphics[width=0.9\linewidth]{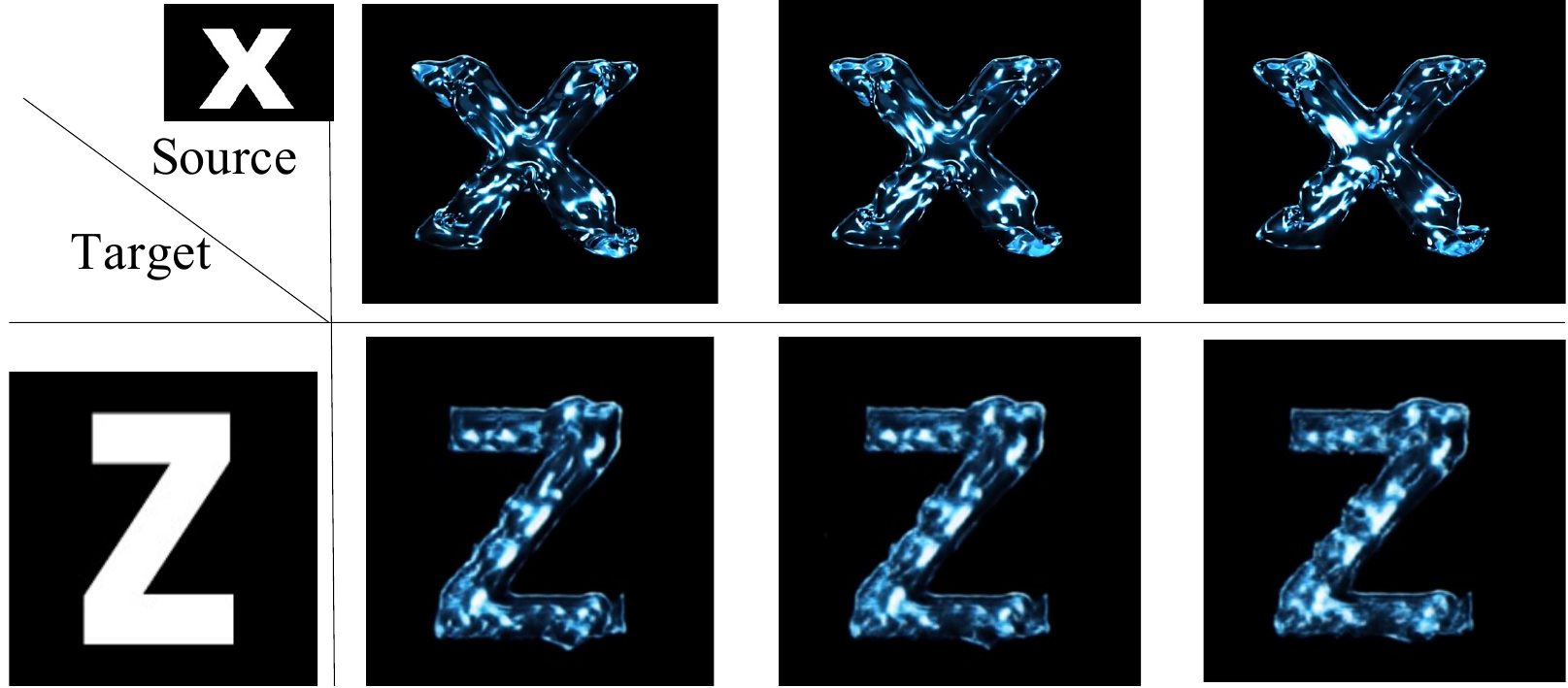}
    \caption{Our results on water dynamic effects transfer.}
    \label{fig:7}
\end{figure}

\begin{figure}[t]
    \centering
    \includegraphics[width=0.9\linewidth]{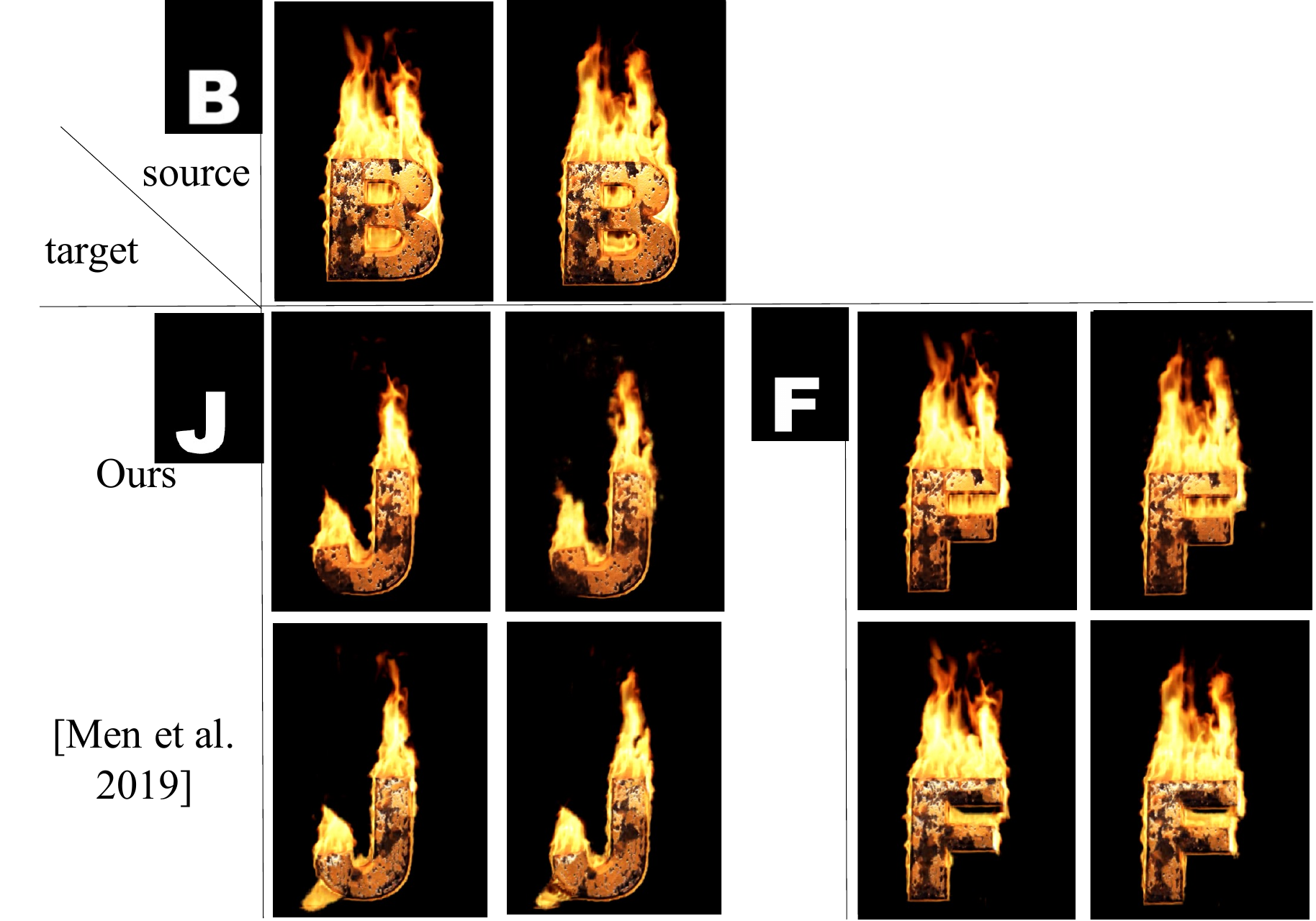}
    \caption{Comparison on moving flame dynamic effects. The exemplar moves to lower right in the driving video. }
    \label{fig:11}
\end{figure}


\subsection{Comparison} 
\label{sec:comparison}

\subsubsection{Qualitative comparison} 

In this subsection, we compare our method with other existing approaches~\cite{fivser2017example}~\cite{kwatra2005texture}~\cite{yang2017awesome}~\cite{men2019dyntypo} for dynamic typography generation. 

\begin{figure}[!t]
    \centering
    \includegraphics[width=1\linewidth]{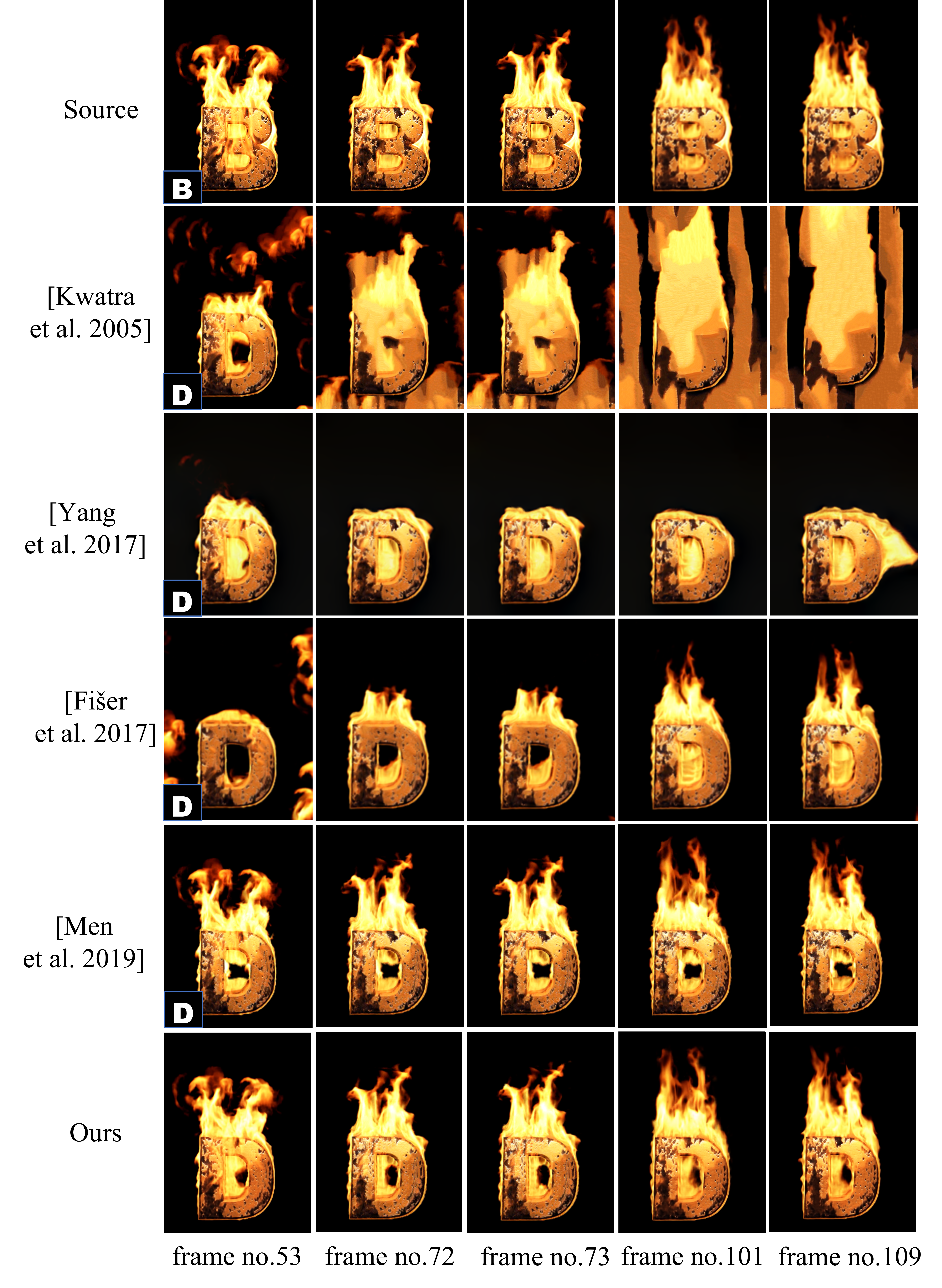}
    \caption{Comparison on flame dynamic effects generated by different methods. Our method synthesizes more natural and visually-pleasing results with the desired text effects while keeping both spatial and temporal consistency.}
    \label{fig:10}
\end{figure}

The synthesis results of some representative frames are depicted in Fig.~\ref{fig:10}. Flow-guided synthesis~\cite{kwatra2005texture} suffers from severe texture distortions and causes ghosting artifacts due to the error accumulation of flows. ~\cite{yang2017awesome} fails to produce stable dynamic effects and synthesizes videos that are temporally inconsistent. ~\cite{fivser2017example} suffers from temporal inconsistency and distinctive artifacts. Distorted textures are inevitably generated by DynTypo~\cite{men2019dyntypo} in some regions due to the side effects brought by the common NNF computed on keyframes. Our method synthesizes more natural and visually-pleasing results with the desired text effects while keeping both spatial and temporal consistency. 

Moreover, Fig.~\ref{fig:11} compares the performance of our method and  DynTypo~\cite{men2019dyntypo} on transferring dynamic and moving flame effects. When the text contains motions and no longer aligns with the original semantic map, the video synthesized by DynTypo contains significant artifacts and inconsistency between frames while our method still provides decent results, indicating its superiority and versatility.

\begin{figure}[!t]
    \centering
    \includegraphics[width=0.5\textwidth]{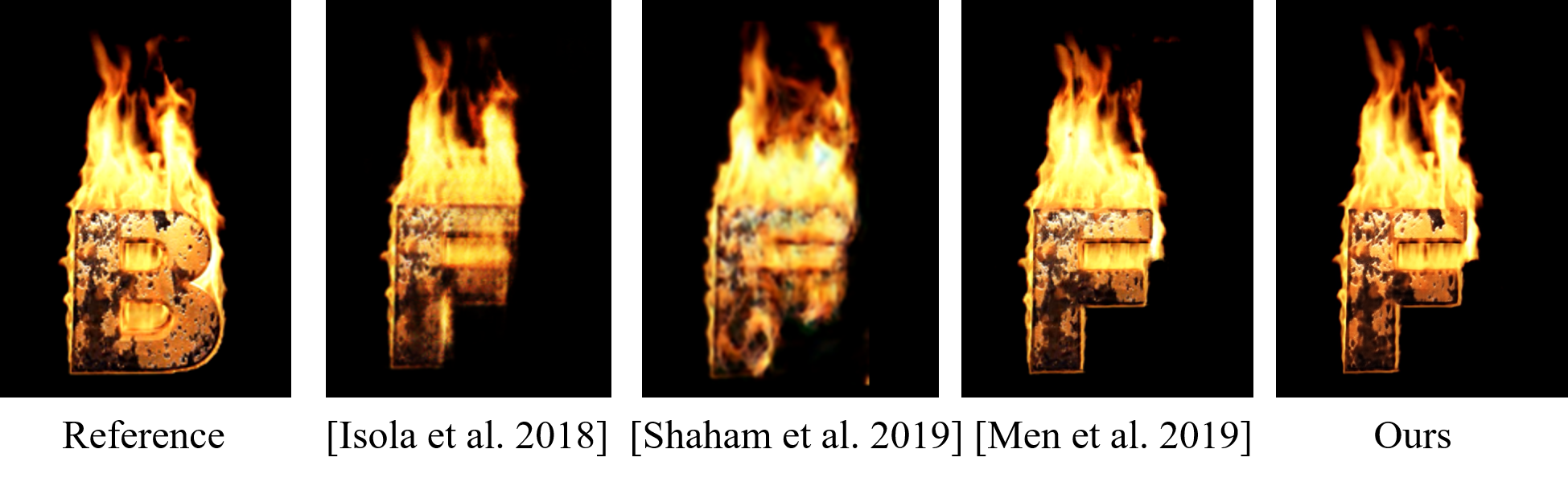}
    \caption{Texture transfer quality comparison.}
    \label{fig:texture_comparison}
\end{figure}

\begin{figure}[!t]
    \centering
    \includegraphics[width=1\linewidth]{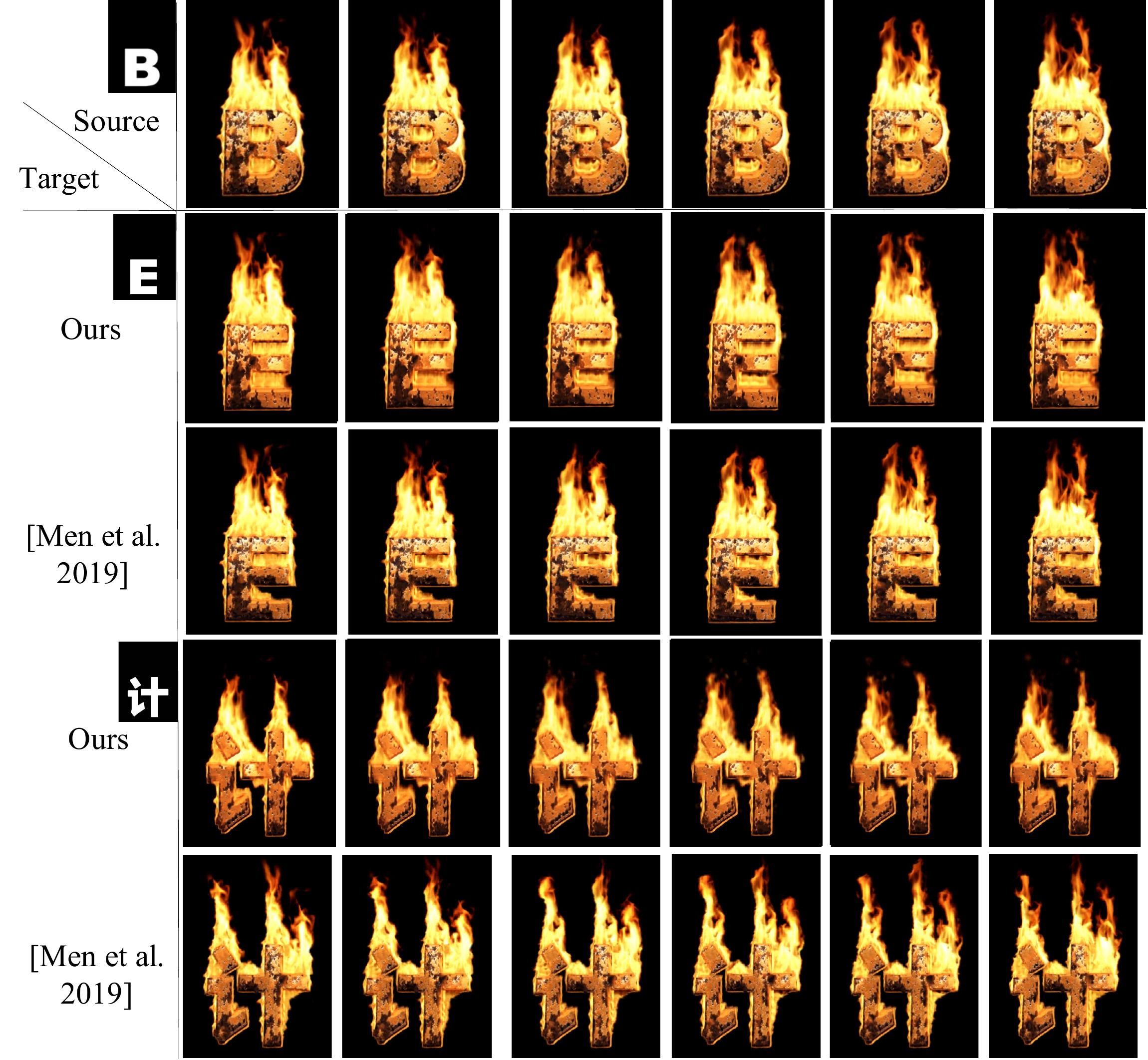}
    \caption{Comparison on flame dynamic effects with Dyntypo. Our DynTexture achieves better spatial consistency of the dynamic effect as well as more natural flame shapes.}
    \label{fig:9}
\end{figure}

As shown in Fig~\ref{fig:texture_comparison}, we provide a detailed comparison for the generated texture quality of our method with other texture transfer methods including Pix2Pix~\cite{Isola_2017_CVPR} (we cut patches to train Pix2Pix for the one-shot learning task), SinGAN~\cite{shaham2019singan} (we change its random variable inputs to semantic map as structure guidance), and DynTypo~\cite{men2019dyntypo}. Fig~\ref{fig:9} shows an extensive comparison between our method and DynTypo~\cite{men2019dyntypo}. Our synthesis results are with the best texture quality and the most natural structure.


\subsubsection{Quantitative comparison} 

 Quantitatively evaluating the results of one-shot learning tasks with no ground truth available is always tough. The similarity metrics, like SSIM, are invalid without ground truth and the perceptual quality metrics, like FID, are also unsuited for one-shot learning tasks including our work, since the data quantity is far too small to obtain reasonable feature distribution. 
 
Due to the fact that our method is implemented based on patches, we found the LPIPS (Learned Perceptual Image Patch Similarity) metric suitable to measure the quality of synthesized videos. Hence, we transfer the flame dynamic texture to several representative types of glyphs (e.g., English and Chinese characters), 30 generated samples in total, and calculate the LPIPS score between the single source video frames and the generated results of~\cite{fivser2017example},~\cite{men2019dyntypo}, and ours. As demonstrated in Table~\ref{table:comparison}, consistent with our visually-pleasing results shown in the paper, our method obtain the best LPIPS score. Moreover, we also conduct a user study where each participant is asked to select the one that is more visually-pleasing from synthesis results generated by our method and (Men et al. 2019)~\cite{men2019dyntypo}. Our experimental results show that, with 64 participants, our method obtains the best preference rate of 79.26\%. 

\begin{table}[t!]
\setlength{\abovecaptionskip}{0.1cm}
\setlength{\belowcaptionskip}{-0.1cm}
\begin{center}
\caption{Quantitative comparison of different methods.}
\begin{tabular*}{0.5\textwidth}{c @{\extracolsep{\fill}} |c|c}
\hline
Model   & LPIPS$\downarrow$     &User Preference$\uparrow$  \\
\hline\hline
Fiser et el. 2017~\cite{fivser2017example}    & 0.51  & -\\
Men et al. 2019~\cite{men2019dyntypo}  &  0.095  & 20.74\% \\
Ours     & \bf 0.085  & \bf 79.26\%  \\
\hline
\end{tabular*}

\label{table:comparison}
\end{center}
\end{table}

\subsection{Other potential applications}
\begin{figure}[t]
    \centering
    \includegraphics[width=0.4\textwidth]{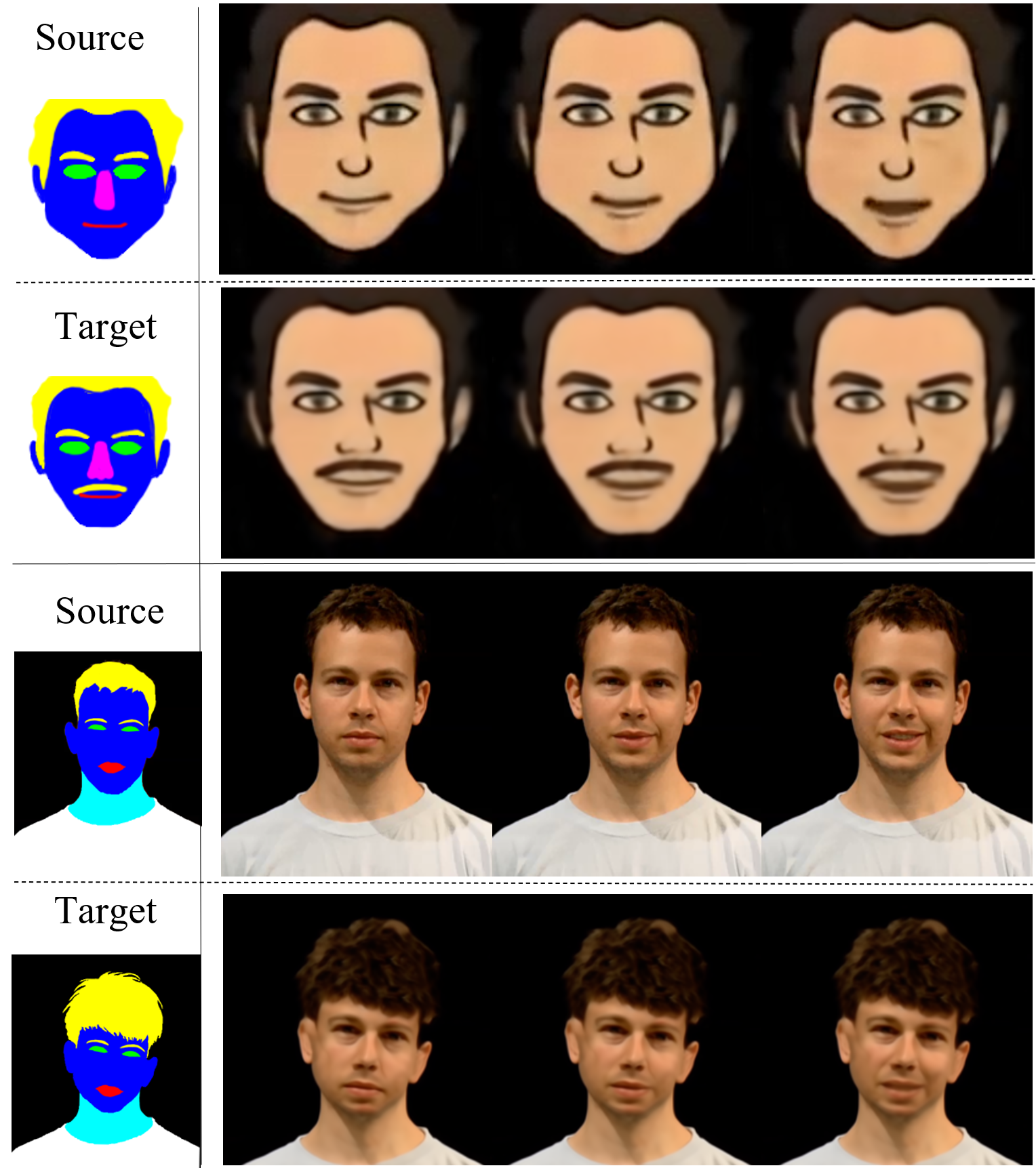}
    \caption{Animation results: cartoons (the first two rows), portraits (the last two rows). Controllable semantic maps serve as structure guidance.}
    \label{fig:8}
\end{figure}

Although the proposed DynTexture is initially designed for dynamic text effects transfer, it does not involve special configurations for typography inputs. Consequently, the proposed patch-cutting and prediction scheme is applicable to various kinds of dynamic effects. For instance, taking semantic maps as the structure guidance, our approach can modify the layout of a source image and animate it according to driving videos. In Fig.~\ref{fig:8}, We demonstrate two examples for layout modification and animation of cartoons and portraits, respectively. In addition, since the proposed deep sequence forecasting module works independently, it can be employed to handle the animation task solely.

\section{Experimental settings and runtime}

We provide details of our experimental settings. Specifically, our patch quantization module is adapted from VQ-VAE. In order to efficiently model small patches, both the encoder and decoder consist of 3 residual convolution blocks, respectively, the number of patch indices is set to 16, the codebook size is set to 256. For subsequent sequence prediction, we used miniGPT with 6 layers and the number of multi-heads for self-attention in each layer is set to 8 (Table~\ref{table:ablation_transformer} shows the choice of the size of the Transformer). The vocabulary size is coherent with the codebook size of VQ-VAE, which is set to 256 in our case. We trained our model with a batch size of 32, using the Adam optimizer whose learning rate is selected as 2.5e-6. The training process takes around 36 hours using two Nvidia GTX2080ti GPUs and the testing takes around 30 minutes per video both depending on the image resolution (patch quantity). 

\begin{table}[t]
\setlength{\abovecaptionskip}{0.1cm}
\setlength{\belowcaptionskip}{-0.1cm}
\begin{center}
\caption{Parameter study of transformer settings. Accuracy scores are the percent ratio of the correctly predicted patches on the validation set of training data. The GPU costs are measured under the batch size of 32 (patches). The runtimes are measured by seconds per epoch during training. We adopt the setting of 6 transformer layers with 8 attention heads for less GPU and runtime cost with high accuracy.}
\begin{tabular*}{0.5\textwidth}{c @{\extracolsep{\fill}} |c|c|c}
\hline
Layers/Heads   & Accuracy$\uparrow$      &GPU cost$\downarrow$ &runtime$\downarrow$ \\
\hline\hline
3/4    & 89.8\%  & 1740 Mib & 257\\
6/8  &  98.6\%  & 3791 Mib & 514\\
12/16     & 99.2\%  & 9533 Mib & 1103\\
\hline
\end{tabular*}
\label{table:ablation_transformer}
\end{center}
\end{table}

\section{Conclusion}
In this paper, we presented a novel model that is able to achieve automatic dynamic texture transfer utilizing both PatchMatch and Transformers. 
To the best of our knowledge, the proposed method is the first neural-based approach to tackle the one-shot dynamic texture transfer problem. 
Experimental results demonstrated the effectiveness and superiority of our method over the state of the art, synthesizing visually-pleasing results for complicated dynamic effects transfer and moving dynamic effects transfer which cannot be properly handled by existing approaches. 
In addition, the proposed patch-cutting and prediction scheme is also applicable to other types of dynamic effects, showing great potentials in motivating researchers working on the one-shot video generation task, as well as inspiring future works in video style transfer and image animation.

\section*{Acknowledgments}
This work was supported by Beijing Nova Program of Science and Technology (Grant No.: Z191100001119077), Center For Chinese Font Design and Research, and Key Laboratory of Science, Technology and Standard in Press Industry (Key Laboratory of Intelligent Press Media Technology)

\bibliographystyle{IEEEtran}
\bibliography{paper}



\begin{IEEEbiography}[{\includegraphics[width=1in,height=1.25in,clip,keepaspectratio]{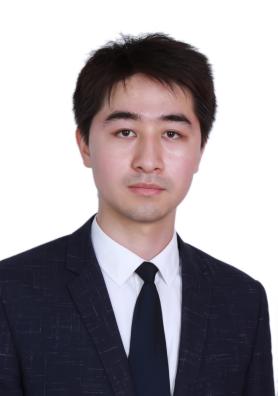}}]{Guo Pu}
received the master's degree from Peking University, China, in 2022. He is currently a Ph.D candidate in Peking University, China, working on his Ph.D thesis at the Wangxuan Institute of Computer Technology, Peking University, China. His research interests include computer vision, image synthesis and texture transfer. He has published papers in related journals and conferences such as TPAMI, SIGGRAPH, CVPR, etc.
\end{IEEEbiography}

\begin{IEEEbiography}[{\includegraphics[width=1in,height=1.25in,clip,keepaspectratio]{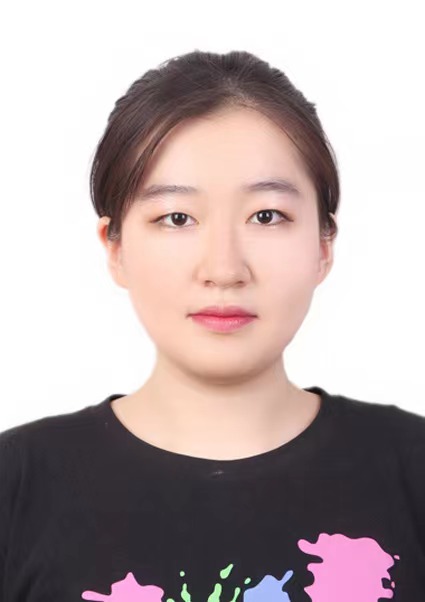}}]{Shiyao Xu}
currently a third-year master student at Wangxuan Institute of Computer Technology, Peking University, China. Her research interests lie in building the bridge between 2D and 3D, specially about image generation and neural rendering.
\end{IEEEbiography}

\begin{IEEEbiography}[{\includegraphics[width=1in,height=1.25in,clip,keepaspectratio]{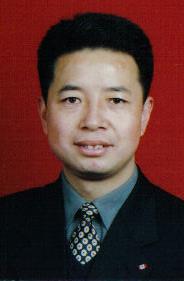}}]{Xixin Cao}
received the Ph.D. degree in computer science in 2000. He is a professor affiliated with School of Software and Microelectronics, Peking University, Beijing, China. His research interests include digital image processing, computer vision and digital multimedia SOC design. He has published more than 50 papers in significant journals and conferences.
\end{IEEEbiography}

\begin{IEEEbiography}[{\includegraphics[width=1in,height=1.25in,clip,keepaspectratio]{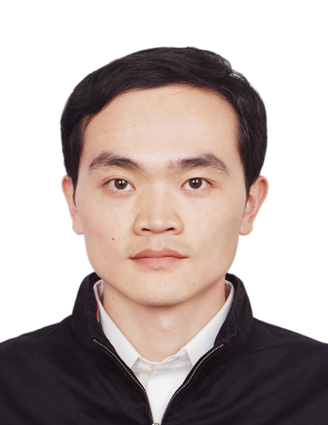}}]{Zhouhui Lian}
received the Ph.D. degree from Beihang University, China, in 2011. He worked as a guest researcher at NIST, Gaithersburg, USA, from 2009 to 2011. He is currently an associate professor at the Wangxuan Institute of Computer Technology, Peking University, China. His main research interests include computer graphics, computer vision, and AI. He has published more than 70 papers in prestigious journals and conferences such as TOG, TPAMI, IJCV, SIGGRAPH/SIGGRAPH Asia, CVPR, etc.
\end{IEEEbiography}

\end{document}